%% file: main.tex
\pdfoutput=1
\documentclass[10pt,twocolumn,letterpaper]{article}
\usepackage{iccv}
\usepackage{times}
\usepackage{epsfig}
\usepackage{graphicx}
\usepackage{amsmath}
\usepackage{amssymb}
\usepackage{ mathrsfs }
\usepackage{xcolor}
\usepackage{tablefootnote}
\usepackage{ stmaryrd }

\usepackage[ruled,vlined,linesnumbered]{algorithm2e}
\usepackage{tabularx} 
\usepackage{multirow}
\usepackage{array}
\usepackage{dblfloatfix}
\newcolumntype{L}[1]{>{\raggedright\let\newline\\\arraybackslash\hspace{0pt}}m{#1}}
\newcolumntype{C}[1]{>{\centering\let\newline\\\arraybackslash\hspace{0pt}}m{#1}}
\newcolumntype{R}[1]{>{\raggedleft\let\newline\\\arraybackslash\hspace{0pt}}m{#1}}
\usepackage{supertabular}
\usepackage{enumitem}
\usepackage{ dsfont }
\usepackage[toc,page]{appendix}

\newcommand{\junk}[1]{}

\usepackage{amsmath}

\usepackage{amssymb}
\usepackage{pifont}
\usepackage[breaklinks=true,bookmarks=false]{hyperref}

 \iccvfinalcopy 

\def\iccvPaperID{2177} 

 \normalsize

\begin{document}

\title{Compact Trilinear Interaction for Visual Question Answering}

\author{Tuong Do$^{1\dagger}$, Thanh-Toan Do$^{2\dagger}$, Huy Tran$^{1}$, Erman Tjiputra$^{1}$, Quang D. Tran$^{1}$\\
{\small $^{1}$AIOZ Pte Ltd, Singapore}
{\small $^{2}$University of Liverpool}}

\maketitle

\input{abstract}

\input{intro}
 
\input{relatedwork}

\input{proposal}

\input{experiment}

\input{conclusion}

{\small
\bibliographystyle{ieee_fullname}
\bibliography{egbib}
}
\end{document}

%% file: abstract.tex
\begin{abstract}
In Visual Question Answering (VQA), answers have a great correlation with question meaning and visual contents. Thus, to selectively utilize image, question and answer information, we propose a novel trilinear interaction model which simultaneously learns high level associations between these three inputs. In addition, to overcome the interaction complexity, we introduce a multimodal tensor-based
PARALIND decomposition which efficiently parameterizes trilinear interaction between the three inputs. Moreover, knowledge distillation is first time applied in Free-form Opened-ended VQA. It is not only for reducing the computational cost and required memory but also for transferring knowledge from trilinear interaction model to bilinear interaction model. The extensive experiments on benchmarking datasets TDIUC, VQA-2.0, and Visual7W show that the proposed compact trilinear interaction model achieves state-of-the-art results when using a single model on all three datasets. 
The source code is available at  \url{https://github.com/aioz-ai/ICCV19_VQA-CTI}.


\end{abstract}

%% file: intro.tex
\section{Introduction}
\let\thefootnote\relax\footnotetext{$\dagger$ \textrm{\ indicates\ equal\ contribution.}}

The aim of VQA is to find out a correct answer for a given question which is consistent with visual content of a given image \cite{malinowski2014towards, VQA, vqav22016}. There are two main variants of VQA which are Free-Form Opened-Ended (FFOE) VQA and Multiple Choice (MC) VQA. In FFOE VQA, an answer is a free-form response to a given image-question pair input, while in MC VQA, an answer is chosen from an answer list for a given image-question pair input.


Traditional approaches  to both VQA tasks mainly aim to learn joint representations between images and questions, while the answers are treated in a ``passive" form, i.e., the answers are only considered as classification targets.  
However, an answer is expected to have high correlation with its corresponding question-image input, hence a jointly and explicitly information extraction from these three inputs will give a highly meaningful joint representation. In this paper, we propose a novel  trilinear interaction model which simultaneously learns high level associations between all three inputs, i.e., image, question, and answer. 




The main difficulty in trilinear interaction is the dimensionality issue which causes expensive computational cost and huge memory requirement. To tackle this challenge, we propose to use PARALIND decomposition \cite{bro2009modelingPARALIND} which factorizes a large tensor into smaller tensors 
which reduces the computational cost and the usage memory.

The proposed trilinear interaction takes images, questions and answers as inputs. However, answer information in FFOE VQA  \cite{2017AgrawalPriorVQA, Teney2017VisualQA, Malinowski2015AskYN, DBLP:conf/cvpr/TeneyLH17} is only available in the training phase but not in the testing phase. To apply the trilinear interaction for FFOE VQA, we propose to use knowledge distillation to transfer knowledge from trilinear model to bilinear model. The distilled bilinear model only requires pairs of image and question as inputs, hence it can be used for the testing phase. 
For MC VQA \cite{zhu2016visual7w, kim2016multimodal, nam2017dual, ilievski2016focused, noh2016image, yu2017multi}, the answer information can be easily extracted, thanks to the given answer list that contains few candidate answers for each image-question pair and is available in both training and testing phases. Thus, the proposed trilinear interaction can be directly applied to MC VQA.

 To evaluate the effectiveness of the proposed model, the extensive experiments are conducted on the benchmarking datasets TDIUC, VQA-2.0, and Visual7W. The results show that the proposed model achieves state-of-the-art results on all three datasets. 
 

The main contributions of the paper are as follows. (i) We propose a novel trilinear interaction model which simultaneously learns high level joint presentation between image, question, and answer information in VQA task. (ii) We utilize PARALIND decomposition to deal with the dimensionality issue in trilinear interaction. (iii) To make the proposed trilinear interaction applicable for FFOE VQA, we propose to use knowledge distillation for transferring knowledge from trilinear interaction model to bilinear interaction model. The remaining of this paper is organized as follows. Section~\ref{sec:related} presents the related work. Section~\ref{Sec:proposal} presents the proposed compact trilinear interaction (CTI). Section~\ref{sec:VQA_integrated} presents the proposed models when applying CTI to FFOE VQA and MC VQA. Section~\ref{Sec:Exp} presents  ablation studies, experimental results and analysis. 


%% file: relatedwork.tex
\section{Related Work}
\label{sec:related}
\textbf{Joint embedding in Visual Question Answering.} 
There are different approaches have been proposed for VQA \cite{Kim2018BilinearAN,Benyounes2017MUTANMT,Fukui2016MultimodalCB, DBLP:conf/iccv/YuY0T17, kim2016hadamard, xu2016ask, Ma2017VisualQA,bottom-up2017, dense-attention, lu2016hierarchical, teney2016zero, zhou2015simple, noh2016training, Teney2017VisualQA}. 
Most of the successful methods focus on learning joint representation between the input question and image \cite{Fukui2016MultimodalCB,Benyounes2017MUTANMT,Kim2018BilinearAN,DBLP:conf/iccv/YuY0T17}. 
In the state-of-the-art VQA, the features of the input image and question are usually represented under matrix forms. E.g., each image is described by a number of interested regions, and each region is represented by a feature vector. Similar idea is applied for question, e.g., an question contains a number of words and each word is represented by a feature vector. A  fully expressive interaction between an image region and a word should be the outer product between their two corresponding vectors~\cite{Fukui2016MultimodalCB}. The outer product allows a multiplicative interaction between all elements of both vectors.  However, a fully bilinear interaction using outer product between every possible pairs of regions and words will dramatically increase the output space. Hence instead of directly computing the fully bilinear with outer product, most of works try to compress or decompose the fully bilinear interaction. 

In~\cite{Fukui2016MultimodalCB}, the authors proposed the Multimodal Compact Bilinear pooling which is an efficient method to compress the bilinear interaction. The method works by projecting the visual  and linguistic features to a higher dimensional space and then convolving both vectors efficiently by using element-wise product in Fast Fourier Transform space.
In~\cite{Benyounes2017MUTANMT}, the authors proposed  Multimodal Tucker Fusion which is a tensor-based Tucker decomposition to efficiently parameterize bilinear interaction between visual and linguistic representations. 
In~\cite{DBLP:conf/iccv/YuY0T17}, the author proposed Factorized Bilinear Pooling that uses two low rank matrices to approximate the fully bilinear interaction. 
Recently, in~\cite{Kim2018BilinearAN} the authors proposed Bilinear Attention Networks (BAN) that finds bilinear attention distributions to utilize given visual-linguistic information seamlessly. BAN also uses low rank approximation to approximate the bilinear interaction for each pair of vectors from image and question.

There are other works that consider answer information, besides image and question information, to improve VQA performance \cite{jabri2016revisiting, gan2017vqs, shih2016look,hu2018learningfPMC,wang2018structuredSTL,schwartz2017high}.
Typically, in \cite{hu2018learningfPMC}, the authors learned two embedding functions to transform an image-question pair and an answer into a joint embedding space. The distance between the joint embedded image-question and the embedded answer is then measured to determine the output answer. 
In \cite{wang2018structuredSTL}, 
the authors computed joint representations between image and question, and between image and answer. They then learned a joint embedding between the two computed representations. 

In \cite{schwartz2017high}, 
the authors computed ``ternary potentials" which capture the dependencies between three inputs, i.e., image, question, and answer. For every triplet of vectors, each from each different input, to compute the interaction between three vectors, instead of calculating the outer products, the author computed the sum of element-wise product of the three vectors. This greatly reduces the computational cost but it might not be expressive enough to fully capture the complex associations between the three vectors.


Different from previous works that mainly aim to learn the joint representations from pairs of modalities~\cite{Fukui2016MultimodalCB,Benyounes2017MUTANMT,Kim2018BilinearAN,DBLP:conf/iccv/YuY0T17,hu2018learningfPMC,wang2018structuredSTL} or greatly simplify the interaction between the three modalities by using the element-wise operator~\cite{schwartz2017high}, in this paper, we propose a principle and direct approach -- a trilinear interaction model, which simultaneously learns a joint representation between three modalities. In particular, we firstly derive a fully trilinear interaction between three modalities. We then rely on a decomposition approach to develop a compact model for the interaction.

\textbf{Knowledge Distillation.} Knowledge Distillation is a general approach for transferring knowledge from a cumbersome model (teacher model) to a lighten model (student model) \cite{hinton2015distilling,gupta2016cross,romero2014fitnets,chen2017learning,ba2014deep}. In FFOE VQA, the trilinear interaction model, which takes image, question, and answer as inputs, can only be applied for training phase but not for testing phase due to the omission of answer in testing. To overcome this challenge and also to reduce computational cost, inspired from the Hinton's seminar work \cite{hinton2015distilling}, we propose to use knowledge distillation to transfer knowledge from trilinear model to bilinear model.

\junk


%% file: proposal.tex
\section{Compact Trilinear Interaction (CTI)}
\label{Sec:proposal}
\subsection{Fully parameterized trilinear interaction}

Let $M = \{M_1, M_2, M_3\}$  be the representations of three inputs. 
$M_t \in \mathds{R}^{n_t \times d_t}$, where $n_t$ is the number of channels of the input $M_t$ and $d_t$ is the dimension of each channel. 
For example, if $M_1$ is the region-based representation for an image, then $n_1$ is the number of regions and $d_1$ is the dimension of the feature representation for each region. Let $m_{t_e} \in \mathds{R}^{1 \times d_{t}}$ be the $e^{th}$ row of $M_t$, i.e., the feature representation of $e^{th}$ channel in $M_t$, where $t \in \{1, 2, 3\}$.


The joint representation resulted from a fully parameterized trilinear interaction over the three inputs is presented by  $z \in \mathds{R}^{d_z}$ which is computed as follows
\begin{equation}
z^T= \left(\left(\mathcal{T} \times_1 vec(M_1)  \right) \times_2 vec(M_2)\right) \times_3 vec(M_3)
\label{eq:hypothesis}
\end{equation}
where $\mathcal{T} \in \mathds{R}^{d_{M_1} \times d_{M_2} \times d_{M_3} \times d_z}$ is a learning tensor; $d_{M_t} = n_t \times d_t$; $vec(M_t)$ is a vectorization of $M_t$ which outputs a row vector; operator $\times_i$ denotes the $i$-mode tensor product. 

The  tensor $\mathcal{T}$ helps to learn the interaction between the three input through $i$-mode product. 
However, learning such a large tensor $\mathcal{T}$ is infeasible when the dimension $d_{M_t}$ of each input modality is high, which is the usual case in VQA. Thus, it is necessary to reduce the size $\mathcal{T}$ to make the learning feasible.


Inspired by \cite{Yang2016StackedAN}, we rely on the idea of \textit{unitary attention} mechanism. Specifically, let $z_p \in \mathds{R}^{d_z}$ be the joint representation of $p^{th}$ triplet of channels where each channel in the triplet is from a different  input. The representation of each channel in a triplet is $m_{1_i}, m_{2_j}, m_{3_k}$, where $i \in  [1,n_1], j \in  [1,n_2], k \in  [1,n_3]$, respectively. There are $n_1 \times n_2 \times n_3$ possible triplets over the three inputs. The joint representation $z_p$ resulted from a fully parameterized trilinear interaction over three channel representations $m_{1_i}, m_{2_j}, m_{3_k}$ of $p^{th}$ triplet is computed as
\begin{equation}
z_p^T= \left(\left(\mathcal{T}_{sc} \times_1 m_{1_i} \right)\times_2 m_{2_j} \right)\times_3 m_{3_k}
\label{eq:triplet_compute}
\end{equation}
where $\mathcal{T}_{sc} \in \mathds{R}^{d_{1} \times d_{2} \times d_{3} \times d_z}$  is the learning tensor between channels in the triplet.

Follow the idea of \textit{unitary attention}~\cite{Yang2016StackedAN}, the joint representation 
$z$ is approximated by using joint representations of all triplets described in (\ref{eq:triplet_compute}) instead of using fully parameterized interaction over three inputs as in (\ref{eq:hypothesis}). Hence, we compute
\begin{equation}
z = \sum_p \mathcal{M}_p z_p
\label{eq:Unitary}
\end{equation}
Note that in (\ref{eq:Unitary}), we compute a weighted sum over all possible triplets. The $p^{th}$ triplet is associated with a scalar weight $\mathcal{M}_p$. The set of $\mathcal{M}_p$ is called as the attention map $\mathcal{M}$, where $\mathcal{M} \in \mathds{R}^{n_1 \times n_2 \times n_3}$.

The attention map $\mathcal{M}$ 
resulted from a reduced parameterized trilinear interaction over three inputs $M_1, M_2$ and $M_3$ is computed as follows
\begin{equation}
\mathcal{M} = \left( \left(\mathcal{T}_\mathcal{M} \times_1 M_1 \right) \times_2 M_2 \right)\times_3 M_3
\label{eq:tri_attmap}
\end{equation}
where $\mathcal{T}_\mathcal{M} \in \mathds{R}^{d_1 \times d_2 \times d_3}$ is the learning tensor of attention map $\mathcal{M}$. 
Note that the learning tensor $\mathcal{T}_\mathcal{M}$ in (\ref{eq:tri_attmap}) has a reduced size compared to the learning tensor $\mathcal{T}$ in (\ref{eq:hypothesis}).





By integrating (\ref{eq:triplet_compute}) into  (\ref{eq:Unitary}), the joint representation $z$ in (\ref{eq:Unitary}) can be   rewritten as
\begin{equation}
z^T= \sum_{i=1}^{n_1}\sum_{j=1}^{n_2}\sum_{k=1}^{n_3} \mathcal{M}_{ijk}\left( \left( \left( \mathcal{T}_{sc} \times_1 m_{1_i}\right) \times_2 m_{2_j} \right) \times_3 m_{3_k} \right)
\label{eq:transformed_hypo}
\end{equation}
where $\mathcal{M}_{ijk}$ in (\ref{eq:transformed_hypo}) is actually a scalar attention weight $\mathcal{M}_p$ of the attention map $\mathcal{M}$ in (\ref{eq:tri_attmap}).


It is also worth noting from (\ref{eq:transformed_hypo}) that to compute $z$, instead of learning the  large tensor $\mathcal{T} \in \mathds{R}^{d_{M_1} \times d_{M_2} \times d_{M_3} \times d_z}$ in (\ref{eq:hypothesis}), we now only need to learn two smaller tensors  $\mathcal{T}_{sc} \in \mathds{R}^{d_{1} \times d_{2} \times d_{3} \times d_z}$  in (\ref{eq:triplet_compute}) and $\mathcal{T}_\mathcal{M} \in \mathds{R}^{d_1 \times d_2 \times d_3}$ in (\ref{eq:tri_attmap}).

\subsection{Parameter factorization}
\label{sub:factorization}

Although the large tensor $\mathcal{T}$ of trilinear interaction model is replaced by two smaller tensors $\mathcal{T}_\mathcal{M}$ and $\mathcal{T}_{sc}$, the dimension of these two tensors still large which makes the learning difficult. 
To further reduce the computational complexity, the PARALIND decomposition \cite{bro2009modelingPARALIND} is applied for $\mathcal{T}_\mathcal{M}$ and $\mathcal{T}_{sc}$.  
The PARALIND decomposition for the learning tensor $\mathcal{T}_\mathcal{M} \in \mathds{R}^{d_1 \times d_2 \times d_3}$ can be calculated as

\begin{equation}
\mathcal{T}_\mathcal{M} \approx \sum^R_{r=1} \left(\left(\mathcal{G}_r \times_1 W_{1_r} \right) \times_2 W_{2_r} \right) \times_3 W_{3_r}
\label{eq:tensorbased_PARALIND}
\end{equation}
\begin{figure}
    \centering
    \includegraphics[width = \columnwidth, keepaspectratio=True]{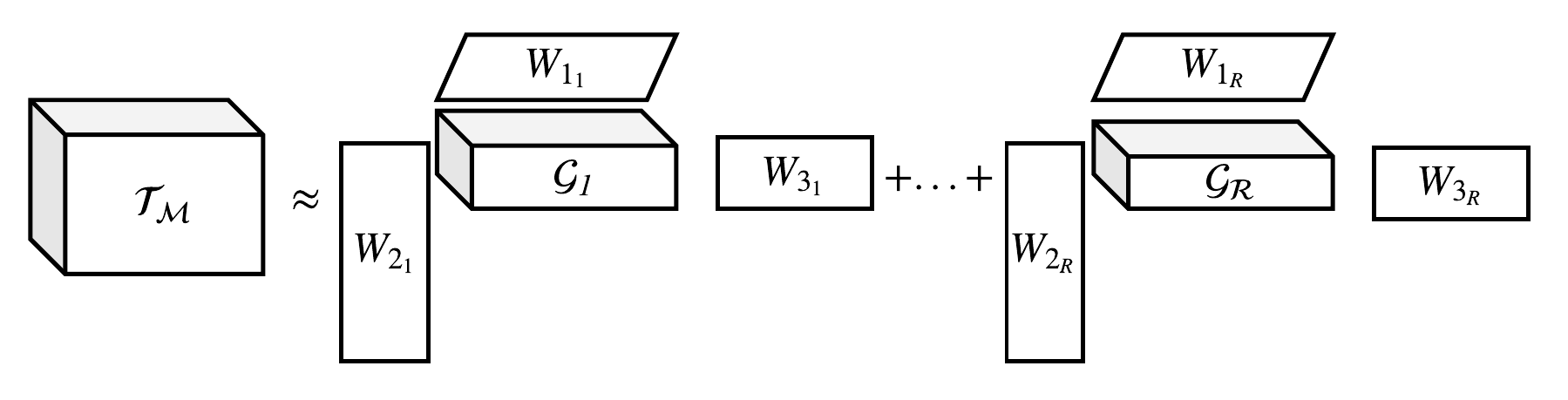}
    \caption{PARALIND decomposition for a tensor $\mathcal{T}_\mathcal{M}$.}
    \label{fig:tensor_based_PARALIND}
    \vspace{0 cm}
\end{figure}

where $R$ is a slicing parameter, establishing a trade-off between the decomposition rate (which is directly related to the usage memory and the computational cost) and the performance.  Each $\mathcal{G}_r \in \mathds{R}^{d_{1_r} \times d_{2_r} \times d_{3_r}}$ is a smaller learnable tensor called Tucker tensor. The number of these Tucker tensors equals to $R$. The maximum value for $R$ is usually set to the greatest common divisor of $d_1, d_2$ and $d_3$. In our experiments, we found that $R=32$ gives a good trade-off between the decomposition rate and the performance.

Here, we have dimension $d_{1_r} = d_1/R$, $d_{2_r} = d_2/R$  and $d_{3_r} = d_3/R$;  
 $W_{1_r} \in \mathds{R}^{d_{1} \times d_{1_r}}$, $W_{2_r} \in \mathds{R}^{d_{2} \times d_{2_r}}$ and  $W_{3_r} \in \mathds{R}^{d_{3} \times d_{3_r}}$ are learnable factor matrices.  
Figure \ref{fig:tensor_based_PARALIND} shows the illustration of PARALIND decomposition for a tensor $\mathcal{T}_\mathcal{M}$.

The shorten form of $\mathcal{T}_\mathcal{M}$ in (\ref{eq:tensorbased_PARALIND}) can be rewritten as
\begin{equation}
\mathcal{T}_\mathcal{M}\approx \sum^R_{r=1} {\llbracket  \mathcal{G}_r; W_{1_r},W_{2_r}, W_{3_r} \rrbracket}
\label{eq:shorten_PARALIND}
\end{equation}

Integrating the learning tensor $\mathcal{T}_\mathcal{M}$ from (\ref{eq:shorten_PARALIND}) into (\ref{eq:tri_attmap}), the attention map $\mathcal{M}$ can be rewritten as
\begin{equation}
\mathcal{M} = \sum^R_{r=1} {\llbracket  \mathcal{G}_r; M_1W_{1_r},  M_2W_{2_r}, M_3W_{3_r}\rrbracket}
\label{eq:tri_attmap_decomp}
\end{equation}

Similar to $\mathcal{T}_\mathcal{M}$, PARALIND decomposition is also  applied  to the tensor $\mathcal{T}_{sc}$ in (\ref{eq:transformed_hypo}) to reduce the complexity. 
It is worth noting that the size of  $\mathcal{T}_{sc}$
 directly effects to the dimension of the joint representation $z \in \mathds{R}^{d_z}$.
Hence, to minimize the loss of information, we set the slicing parameter $R = 1$ and the projection dimension of factor matrices at $d_z$, i.e., the same dimension of the joint representation $z$.

Therefore,  $\mathcal{T}_{sc} \in \mathds{R}^{d_1 \times d_2 \times d_3 \times d_z}$ in (\ref{eq:transformed_hypo}) can be  calculated as
\begin{equation}
\mathcal{T}_{sc} \approx \left(\left(\mathcal{G}_{sc} \times_1 W_{z_1} \right) \times_2 W_{z_2} \right) \times_3 W_{z_3}
\label{eq:T_sc_PARALIND}
\end{equation}
 where $W_{z_1} \in \mathds{R}^{d_{1} \times d_z}$, $W_{z_2} \in \mathds{R}^{d_{2} \times d_z}$, $W_{z_3} \in \mathds{R}^{d_{3} \times d_z}$ are learnable factor matrices and $\mathcal{G}_{sc} \in \mathds{R}^{d_z \times d_z \times d_z \times d_z}$ is a smaller  tensor (compared to $\mathcal{T}_{sc}$).

 Up to now, we already have $\mathcal{M}$ by (\ref{eq:tri_attmap_decomp}) and $\mathcal{T}_{sc}$ by (\ref{eq:T_sc_PARALIND}), hence, we can compute $z$ using (\ref{eq:transformed_hypo}).
 $z$ from (\ref{eq:transformed_hypo}) can be rewritten as 
 \begin{equation}
\begin{aligned}
 &z^T= \sum_{i=1}^{n_1}\sum_{j=1}^{n_2}\sum_{k=1}^{n_3}\\
 & \mathcal{M}_{ijk}\left( \left(\mathcal{G}_{sc} \times_1 m_{1_i}W_{z_1}\right) \times_2 m_{2_j}W_{z_2}\right) \times_3 m_{3_k}W_{z_3}
\end{aligned}
\label{eq:z1}
\end{equation}
Here, it is interesting to note that $\mathcal{G}_{sc} \in \mathds{R}^{d_z \times d_z \times d_z \times d_z}$ in (\ref{eq:z1}) has rank $1$. Thus, the result got from $i$-mode tensor products in (\ref{eq:z1})  can be approximated by the Hadamard products without the presence of rank-1 tensor $\mathcal{G}_{sc}$ \cite{kolda2009tensor}.
In particular, $z$ in (\ref{eq:z1}) can be computed without using $\mathcal{G}_{sc}$ as
 \begin{equation}
z^T = \sum_{i=1}^{n_1}\sum_{j=1}^{n_2}\sum_{k=1}^{n_3} \mathcal{M}_{ijk} \left( m_{1_{i}}W_{z_1} \circ m_{2_{j}}W_{z_2} \circ m_{3_{k}}W_{z_3}\right)
\label{eq:final_hypo}
\end{equation}
Note that $d_z$, which is the joint embedding dimension, is a user-defined parameter which makes a trade-off between the capability of the representation and the computational cost. In our experiments, we found that $d_z = 1,024$ gives a good trade-off. 

\section{Compact Trilinear Interaction for VQA}
\label{sec:VQA_integrated}

The input for training VQA is  set of $(V,Q,A)$ in which $V$ is an image representation; $V$ $ \in \mathds{R}^{v \times d_v}$ where $v$ is the number of interested regions (or bounding boxes) in the image and $d_v$ is the dimension of the representation for a region; $Q$ is a question representation; $Q \in \mathds{R}^{q \times d_q }$ 
where $q$ is the number of hidden states and $d_q$ is the dimension for each hidden state. 
$A$ is an answer representation; $A \in \mathds{R}^{a \times d_a}$ 
where $a$ is the number of hidden states and $d_a$ is the dimension for each hidden state. 

By applying the Compact Trilinear Interaction (CTI) to each $(V, Q, A)$, we achieve the joint representation $z \in \mathds{R}^{d_z}$. 
Specifically, we firstly compute the  attention map $\mathcal{M}$ by (\ref{eq:tri_attmap_decomp}) as follows
\begin{equation}
\mathcal{M} = \sum^R_{r=1} {\llbracket  \mathcal{G}_r; V W_{v_r},  Q W_{q_r}, A W_{a_r}\rrbracket}
\label{eq:tri_attmap_decomp_VQA}
\end{equation}

Then the joint representation $z$ is computed by  (\ref{eq:final_hypo}) as follows
\begin{equation}
z^T= \sum_{i=1}^{v}\sum_{j=1}^{q}\sum_{k=1}^{a} \mathcal{M}_{ijk}\left( V_{i}W_{z_v} \circ Q_{j}W_{z_q} \circ A_{k}W_{z_a}\right)
\label{eq:final_hypo_VQA}
\end{equation}
where $W_{v_r},W_{q_r}, W_{a_r}$ in (\ref{eq:tri_attmap_decomp_VQA}) and $W_{z_v},W_{z_q}, W_{z_a}$ in (\ref{eq:final_hypo_VQA}) are learnable factor matrices; each $\mathcal{G}_r$ in (\ref{eq:tri_attmap_decomp_VQA}) is a learnable Tucker tensor.

\subsection{Multiple Choice Visual Question Answering}
\label{Sub:MC}

To make a fair comparison to the state of the art in MC VQA \cite{hu2018learningfPMC,wang2018structuredSTL}, we follow the representations used in those works. Specifically, each input question and each answer are trimmed to a maximum of 12 words which will then be zero-padded if shorter than 12 words. Each word is then represented by a 300-D GloVe word embedding \cite{pennington2014glove}. Each image is represented by a $14 \times 14 \times 2048$ grid feature (i.e., $196$ cells; each cell is with a $2,048$-D feature), extracted from the second last layer of ResNet-152 which is pre-trained on ImageNet \cite{he2016deepResnet152}.

\begin{figure*}
    \centering
    \includegraphics[width = 14 cm, height = 8.5 cm ,keepaspectratio=False]{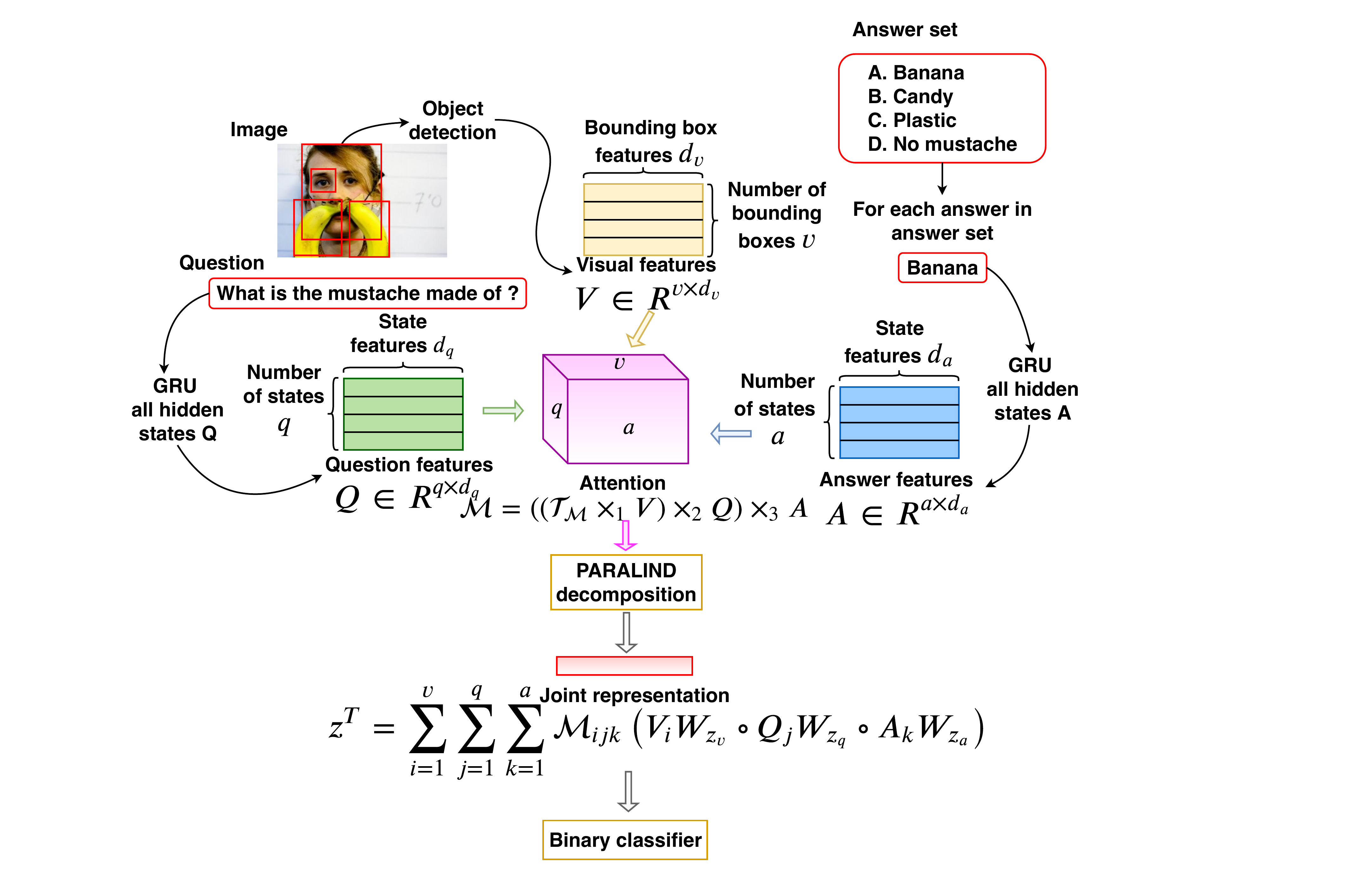}
    \caption{The proposed model when CTI is applied to MC VQA. The details are described in Section \ref{Sub:MC}. Best view in color.}
    \label{fig:MC}
    \vspace{-0.25 cm}
\end{figure*}

Follow \cite{wang2018structuredSTL}, input samples are divided into positive samples and negative samples. A positive sample, which is labelled as $1$ in binary classification, contains image, question and the right answer. A negative sample, which is labelled as $0$ in binary classification, contains image, question, and the wrong answer.
These samples are then passed through our proposed CTI to get the joint representation $z$. The joint representation is passed through a  binary classifier to get the prediction. The Binary Cross Entropy loss is used for training the proposed model. Figure \ref{fig:MC} visualizes the proposed model  when applying CTI to MC VQA.

\subsection{Free-Form Opened-Ended Visual Question Answering}
\label{sub:FFOE}

\begin{figure*}
    \centering
    \includegraphics[width = 15 cm, height = 10 cm ,keepaspectratio=False]{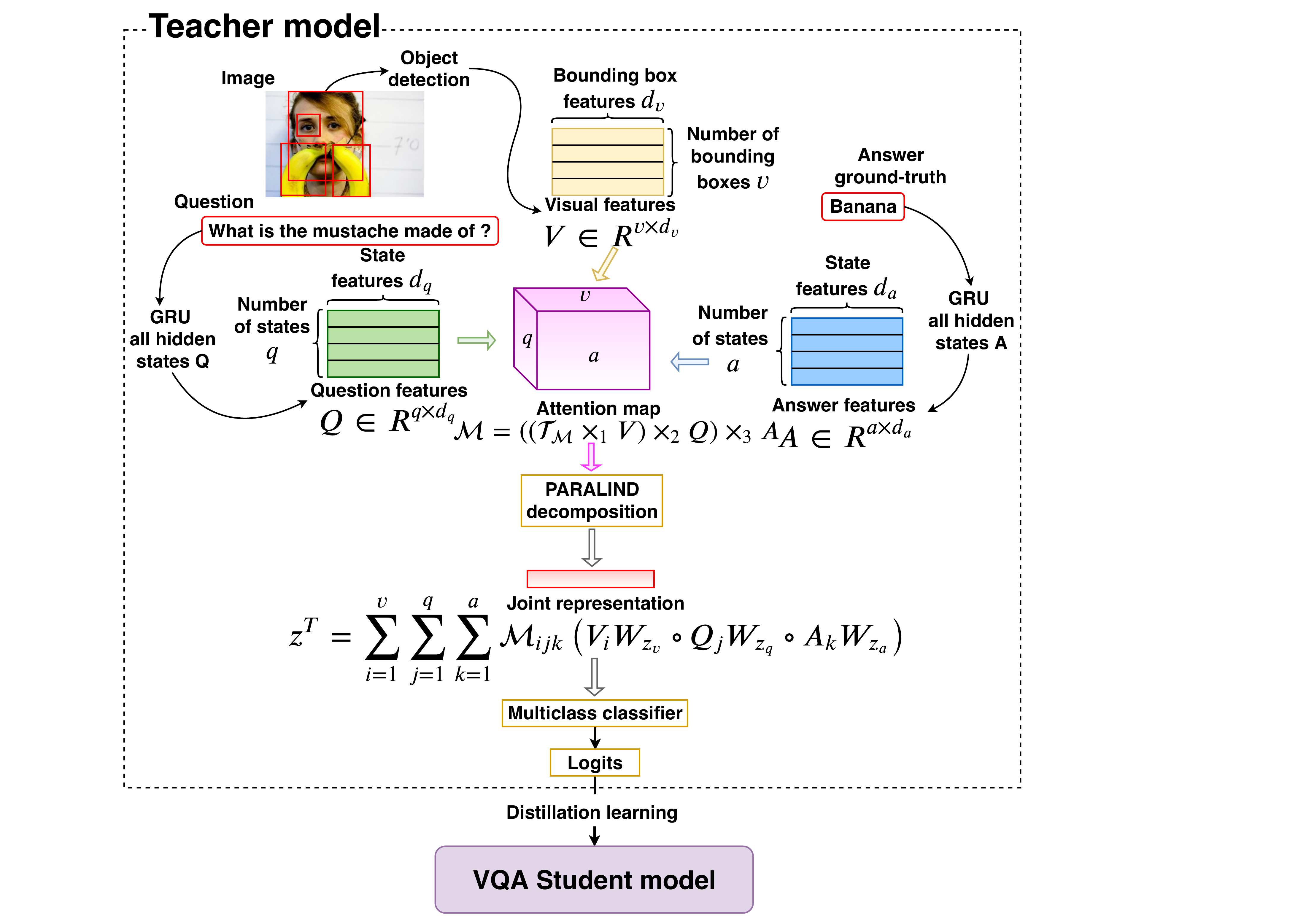}
    \caption{The proposed model when CTI is applied on FFOE VQA. The details are described in Section \ref{sub:FFOE}. Best view in color.}
    \label{fig:FFOE}
    \vspace{-0.25 cm}
\end{figure*}

Unlike MC VQA, FFOE VQA treats the answering as a classification problem over the set of predefined answers. Hence the set possible answers for each question-image pair is much more than the case of MC VQA. Therefore the model design proposed in Section \ref{Sub:MC}, i.e. for each question-image input, the model takes every possible answers from its answer list to computed the joint representation, causes high computational cost. 
In addition, the proposed CTI requires all three $V, Q, A$ inputs to compute the joint representation. However, during the testing, 
there are no available answer information in FFOE VQA. 
To overcome these challenges, we propose to use Knowledge Distillation \cite{hinton2015distilling} to transfer the learned knowledge from a teacher model to a  student model. Figure \ref{fig:FFOE} visualizes the proposed design for FFOE VQA. 

Our teacher model takes triplets of \textit{image-question-right answer} as inputs. Each triplet is passed through the proposed CTI to get the joint representation $z$. The joint representation $z$ is then passed through a multiclass classifier (over the set of predefined answers) to get the prediction which is similar to \cite{tip-trick}. The Cross Entropy loss is used for training the teacher model. 
Regarding the student models, any state-of-the-art VQA can be used. In our experiments, we use BAN2 \cite{Kim2018BilinearAN} or SAN \cite{Yang2016StackedAN} as student models. The student models take pairs of image-question as inputs and treat the prediction as a mutilclass classification problem. 
The loss function for the student model is defined as

\begin{equation}
\mathcal{L}_{KD} = \alpha T^2 \mathcal{L}_{CE}(Q^\tau_S, Q^\tau_T) + (1-\alpha)\mathcal{L}_{CE}(Q_S,y_{true})
\label{eq:distil_loss}
\end{equation}
where $\mathcal{L}_{CE}$ stands for Cross Entropy loss; $Q_S$ is the standard softmax output of the student; $y_{true}$ is the ground-truth answer labels;
$\alpha$ is a  hyper-parameter for controlling the importance of each loss component; $Q^\tau_S, Q^\tau_T$ are the softened outputs  of the student and the teacher using the same temperature parameter $T$ \cite{hinton2015distilling}, which are computed as follows
\begin{equation}
Q^\tau_i = \frac{exp(l_i/T)}{\sum_{i} exp(l_i/T)}
\label{eq:convert_softmax_loss}
\end{equation}

where for both teacher and the student models, the logit $l$ is the predictions outputted by the corresponding classifiers. 

Following by the current state of the art in FFOE VQA \cite{Kim2018BilinearAN}, for image representation, we use object detection-based features with FPN detector (ResNet152 backbone)\cite{2016LinFPN}, in which the number of maximum detected bounding boxes is set to $50$. For question and answer representations, we trim question and answer to a maximum of 12 words which will then be zero-padded if shorter than 12 words. Each word is then represented by a 600-D vector that is a concatenation of the 300-D GloVe word embedding \cite{pennington2014glove} and the augmenting embedding from training data as~\cite{Kim2018BilinearAN}. In the other words, a question is with a representation with size $12 \times 600$. It is similar for answer.

%% file: experiment.tex
\section{Experiments}
\label{Sec:Exp}
\subsection{Dataset and evaluation protocol}
\label{Sub:Exp_dataset}
\textbf{Dataset.} We conduct the experiments on three benchmarking VQA datasets that are Visual7W \cite{zhu2016visual7w} for the MC VQA,  VQA-2.0~\cite{vqav22016} and TDIUC~\cite{Kushal2018Tdiuc} for the FFOE VQA. 
We use training set to train and validation set to evaluate in all mentioned datasets when conducting ablation study.


\textbf{Implementation details.}
\label{subsec:implement}
Our CTI is implemented using PyTorch \cite{paszke2017automaticPyTorch}. The experiments are conducted on a NVIDIA Titan V GPUs with 12GB RAM. 
In all experiments, the learning rate is set to $10^{-3}$.  
Batch size is set to $128$ for training MC VQA and $256$ for training FFOE VQA. 
When training both MC VQA model (Section \ref{Sub:MC}) and FFOE VQA model (Section \ref{sub:FFOE}), except the image representation extraction, other components are trained end-to-end.
The temperature parameter $T$ in (\ref{eq:convert_softmax_loss}) is set to $3$.  
The dimension of the joint representation $z$ is set at $1,024$ for both MC VQA and FFOE VQA.

\textbf{Evaluation Metrics.} We follow the literature~\cite{VQA,Kushal2018Tdiuc, zhu2016visual7w} in which the evaluation metrics for each VQA task 
are different. For FFOE VQA, the single accuracy, which is a standard VQA accuracy (\textit{Acc}) \cite{VQA}, is applied for both TDIUC and VQA-2.0 datasets. 
In addition, due to the imbalance in the question types of TDIUC dataset, follow \cite{Kushal2018Tdiuc}, we also report four other metrics that compensate for the skewed question-type distribution. 
They are Arithmetic MPT (\textit{Ari}), Arithmetic Norm-MPT (\textit{Ari-N}), Harmonic MPT (\textit{Har}), and Harmonic Norm-MPT (\textit{Har-N}). 
For MC VQA, we follow the evaluation metric (\textit{Acc-MC}) proposed by \cite{zhu2016visual7w} in which the performance is measured by the portion of correct answers selected by the  VQA model from the candidate answer set. 

\subsection{Ablation study}
\label{Sub:alb}

\begin{table}[!t]
\begin{center}
\small
\begin{tabular}{c|c|c|c|c|c|c}
\hline
\multirow{2}{*}{\textbf{QT}}    & \multirow{2}{*}{\textbf{Models}} & \multicolumn{5}{c}{\textbf{Evaluation metrics}}                                \\ \cline{3-7} 
                                                                                   &                                  & \textbf{Acc}  & \textbf{Ari}  & \textbf{Har}  & \textbf{Ari-N} & \textbf{Har-N} \\ \hline
\multirow{4}{*}{\textbf{\begin{tabular}[c]{@{}c@{}}with\\ Abs\end{tabular}}}    & BAN2-CTI                           & \textbf{87.0} & \textbf{72.5} & \textbf{65.5} & \textbf{45.8}  & \textbf{28.6}  \\ 
                                                                                   & BAN2\cite{Kim2018BilinearAN}                          & 85.5          & 67.4          & 54.9          & 37.4           & 15.7           \\  
                                                                                   & SAN-CTI                           & 84.5          & 68.7          & 59.9          & 41.3           & 23.3           \\  
                                                                                   & SAN\cite{Yang2016StackedAN}                        & 82.3          & 65.0          & 53.7          & 35.4           & 14.7           \\ \hline
\multirow{4}{*}{\textbf{\begin{tabular}[c]{@{}c@{}}w/o\\ Abs\end{tabular}}} & BAN2-CTI                           & \textbf{85.0} & \textbf{70.6} & \textbf{63.8} & \textbf{41.5}  & \textbf{26.9}  \\ 
                                                                                   & BAN2\cite{Kim2018BilinearAN}                           & 81.9          & 64.6          & 52.8          & 31.9           & 14.6           \\
                                                                                   & SAN-CTI                           & 82.8          & 66.7          & 58.1          & 36.8           & 21.8           \\ 
                                                                                   & SAN\cite{Yang2016StackedAN}                            & 79.1          & 62.4          & 51.7          & 30.2           & 13.7           \\ \hline
\end{tabular}
\end{center}
\caption{Overall performance of the proposal and the baselines BAN2, SAN in different evaluation metrics on TDIUC validation set. The performance is shown with and without considering \textit{Absurd} question category. \textbf{BAN2-CTI} and \textbf{SAN-CTI} are student models trained under our proposed CTI teacher model.}
\label{tab:abl_metric_compare}
\vspace{-0.5 cm}
\end{table}

\textbf{The effectiveness of CTI on FFOE VQA.} 
We compare our distilled BAN2 (BAN2-CTI) and distilled SAN (SAN-CTI) student models to the state-of-the-art baselines BAN2 \cite{Kim2018BilinearAN} and SAN \cite{Yang2016StackedAN}. 
Table \ref{tab:abl_metric_compare} presents a comprehensive evaluation on five different metrics on TDIUC. Among all metrics, on overall, our BAN2-CTI and SAN-CTI outperform corresponding baselines  by a noticeable margin. These results confirm the effectiveness of our proposed CTI for learning the joint representation. In addition, the proposed teacher model (Figure \ref{fig:FFOE}) is also effective. It successfully transfers useful learned knowledge to the student models. 
Note that in Table  \ref{tab:abl_metric_compare}, the ``Absurd" question category indicates the cases in which input questions are irrelevant to the image contents. Thus, the answers are always ``does not apply", i.e., ``no answer". Using these meaningless answers when training the teacher causes negative effect when learning the joint representation, hence, reducing the model capacity. 
If the ``Absurd" category is not taken into account, the proposed model achieves more improvements over  baselines. 


\begin{table}[!t]
\begin{center}
\small
\begin{tabular}{l|cc|cc}
\hline
\textbf{Question-types} & \textbf{BAN2-CTI} & \textbf{\begin{tabular}[c]{@{}c@{}}BAN2 \\ \cite{Kim2018BilinearAN}\end{tabular}} & \textbf{SAN-CTI} & \textbf{\begin{tabular}[c]{@{}c@{}}SAN \\ \cite{Yang2016StackedAN}\end{tabular}}  \\ \hline
Scene Rec               & \textbf{94.5}   & 93.1           & \textbf{93.6}   & 92.3           \\ 
Sport Rec               & \textbf{96.3}   & 95.7           & \textbf{95.5}   & \textbf{95.5}  \\ 
Color Attr              & \textbf{74.3}   & 67.5           & \textbf{70.9}   & 60.9           \\ 
Other Attr              & \textbf{60.5}   & 53.2           & \textbf{56.4}   & 46.2           \\ 
Activity Rec            & \textbf{63.2}   & 54.0           & \textbf{54.5}   & 51.4           \\ 
Positional Rec          & \textbf{40.5}   & 27.9           & \textbf{34.3}   & 27.9           \\ 
Sub-Obj Rec             & \textbf{89.3}   & 87.5           & \textbf{87.6}   & 87.5           \\ 
\textbf{Absurd}         & 93.9            & \textbf{98.2}  & 90.6            & \textbf{93.4}  \\ 
Util \& Aff             & \textbf{36.3}   & 24.0           & \textbf{31.0}   & 26.3           \\ 
Obj Pres                & \textbf{96.1}   & 95.1           & \textbf{94.9}   & 92.4           \\ 
Count                   & \textbf{59.7}   & 53.9           & \textbf{55.6}   & 52.1           \\ 
Sentiment               & \textbf{66.1}   & 58.7           & \textbf{59.9}   & 53.6           \\ \hline
\end{tabular}
\end{center}
\caption{Performance (\textit{Acc}) of the proposal and the baselines BAN2, SAN for each question-type on TDIUC validation set. \textbf{BAN2-CTI} and \textbf{SAN-CTI} are student models trained under our compact trilinear interaction teacher model.}
\label{tab:abl_qt}
\vspace{-0.3 cm}
\end{table}

Table \ref{tab:abl_qt} presents detail performances with \textit{Acc} metric over each question category of TDIUC when all categories, including ``Absurd", are used for training.  The results show that we achieve the best results on all question categories but ``Absurd". 
We note that in the real applications, the ``Absurd" question problem may be mitigated in some cases by using a simple trick, i.e., asking a ``presence question" before asking the main question, e.g., we have an image with no human but the main question is ``Is the people wearing hat?", i.e., a ``Absurd" question. By asking a ``presence question" as ``Are there any people in the picture?", we can have a confirmation about the presence of human in the considered image, 
before asking the main question.

\begin{table}[!t]
\begin{center}
\small
\begin{tabular}{|c|c|c|}
\hline
\textbf{\begin{tabular}[c]{@{}c@{}}Ref \\ models\end{tabular}} & \textbf{\begin{tabular}[c]{@{}c@{}}Validation\\ Accuracy\end{tabular}} & \textbf{\begin{tabular}[c]{@{}c@{}}Test-dev\\ Accuracy\end{tabular}} \\ \hline
Bottom-up \cite{tip-trick}                   & 63.2                                                                   & 65.4                                                                \\ \hline
SAN \cite{Yang2016StackedAN}                 & 61.7                                                                   & 63.0                                                                \\ \hline
SAN-CTI                                                       & 62.1                                                                   &63.4                                                                     \\ \hline
BAN2 \cite{Kim2018BilinearAN}                 & 65.6                                                                   & 66.5                                                                \\ \hline
\textbf{BAN2-CTI}                                              & \textbf{66.0}                                                          & \textbf{67.4}                                                       \\ \hline
\end{tabular}
\end{center}
\caption{Performance of the proposal and baselines BAN2, SAN in VQA-2.0 validation set and test-dev set. \textbf{BAN2-CTI} and \textbf{SAN-CTI} are student models trained under proposed teacher model.}
\label{tab:abl_vqa2}
\vspace{0 cm}
\end{table}

Table \ref{tab:abl_vqa2} presents comparative results between our distilled student models and two baselines BAN2, SAN on \textit{Acc} metric on VQA-2.0. Although our proposal outperforms the baselines, the improvement gap is not much. This is understandable because the VQA-2.0 dataset has a large number of questions of which answers are ``yes/no" or contain only one word (i.e., answers for ``number" question types). These answers have little semantic meanings which prevent proposed trilinear interaction from promoting its efficiency.

\textbf{The effectiveness of CTI on MC VQA.} We still use the state-of-the-art BAN2 \cite{Kim2018BilinearAN} and SAN \cite{Yang2016StackedAN} as baselines and conduct experiments on Visual7W dataset. In MC VQA, in both training and testing, each image-question pair has a corresponding answer list that contains four answers. 
To make a fair comparison, we try different pair combinations over three modalities (image, question, and answer) for the baselines BAN2 and SAN. 
Similar to \cite{wang2018structuredSTL}, we find the following combination gives best results for the baselines. Using BAN2 (or SAN), we first compute the joint  representation between image and question; and the joint representation between image and answer. Then, we concatenate the two computed representations to get the joint ``image-question-answer"  representation, and pass it through VQA classifier with cross entropy loss for training the baseline.


\begin{table}[!t]
\begin{center}
\small
\begin{tabular}{c|c|c}
\hline
\multirow{2}{*}{\textbf{\begin{tabular}[c]{@{}c@{}}Ref\\ models\end{tabular}}} & \multicolumn{2}{c}{\textbf{Visual7W validation set}}                                                \\ \cline{2-3} 
                                           & \multicolumn{1}{l|}{\textbf{Acc-MC}} & \multicolumn{1}{l}{\textbf{Number of parameters}} \\ \hline

\textbf{BAN2 \cite{Kim2018BilinearAN}}                             & 65.7                                         & $\sim$ 86.5M                                   \\
\textbf{SAN \cite{Yang2016StackedAN}}                             & 59.3                                         & $\sim$ 69.7M                                   \\ \hline
\textbf{CTI}                               & \textbf{67.0}                                         & $\sim$ 66.5M                                   \\ \hline
\end{tabular}
\end{center}
\caption{The performance (\textit{Acc-MC}) and the number of parameters of the proposed MC VQA model and the baselines BAN2, SAN on Visual7W validation set.}
\label{tab:abl_v7w}
\vspace{-0.3 cm}
\end{table}

\begin{figure}
    \centering
    \includegraphics[width = \columnwidth, keepaspectratio=False]{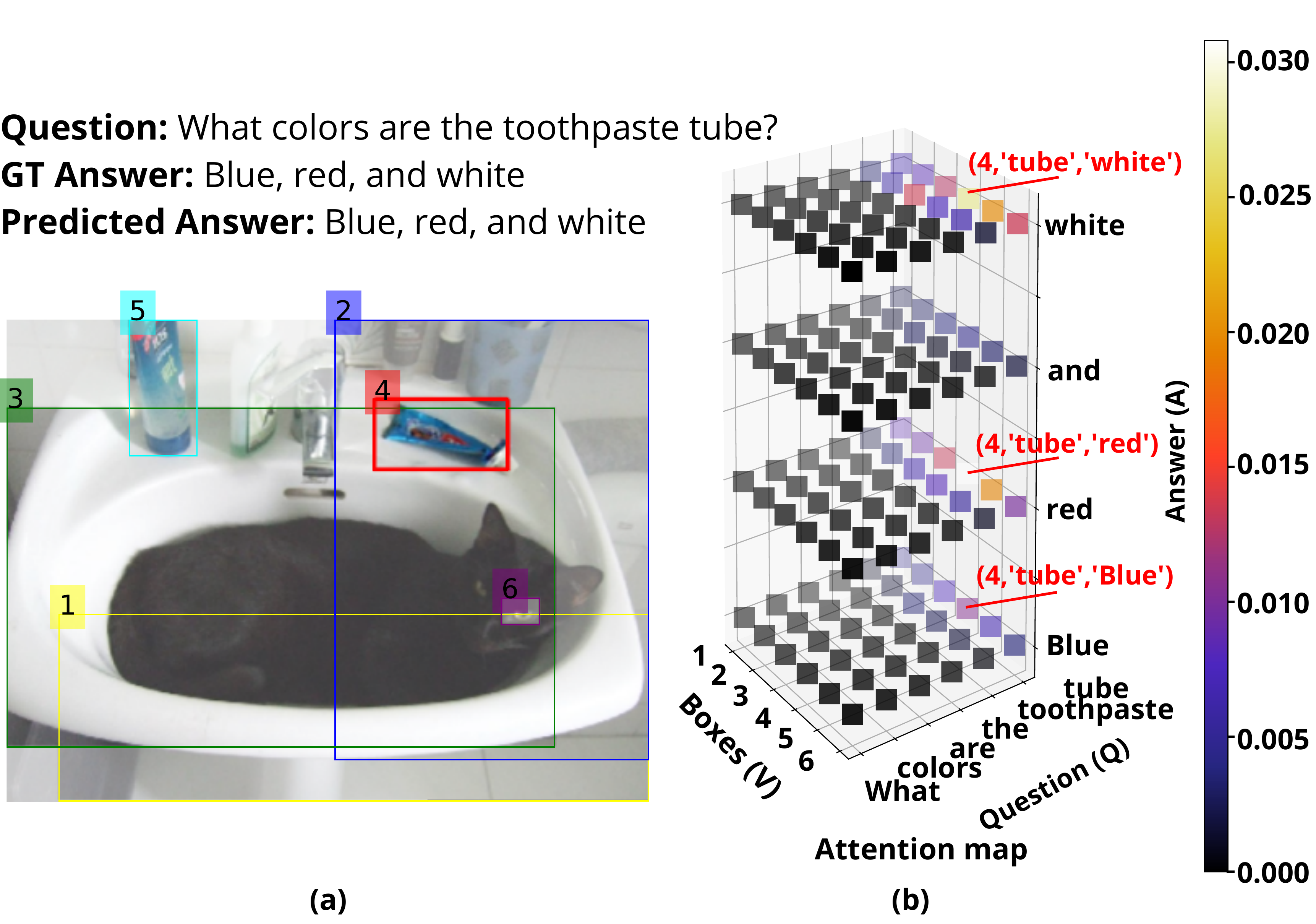}
    \caption{The visualization of an attention map (b) computed from Eq. (\ref{eq:tri_attmap_decomp_VQA}) for an image-question-answer input (a). The attention map indicates attention weights over triplets of \textit{``detected bounding box - word in question - word in answer"}. The higher weight of a triplet is, the more contribution it makes to the joint representation. We can see that three triplets (V=4, Q=``tube", A=``white"), (V=4, Q=``tube", A=``red"), (V=4, Q=``tube", A=``blue") have high weight values. That means that these triplets give high contribution to the joint representation. The input sample (a) is from Visual7W validation set. Best view in color.} 
    \label{fig:Vis}
\vspace{-0.25 cm}
\end{figure}

Table \ref{tab:abl_v7w} presents comparative results on  Visual7W  with \textit{Acc-MC} metric. The results show that our proposed model outperforms the 
baselines by a noticeable margin. These results confirm  that the joint representation learned by the proposed trilinear interaction achieves better performance than the combination of joint representations computed by BAN (or SAN) of pairs of modalities.
In addition, in Table \ref{tab:abl_v7w} we also provide the number of total parameters of our proposed MC VQA model with CTI (Figure~\ref{fig:MC}) and BAN2, SAN. The results show that our model requires less memory than those baselines. That means that the proposed MC VQA model with CTI not only outperforms the baselines in term of accuracy, but also more efficient than those baselines in term of the usage memory. 
Figure \ref{fig:Vis}
visualizes the attention map resulted by CTI for an example of image-question-answer. The attention map is computed by (\ref{eq:tri_attmap_decomp_VQA}). 

\subsection{Comparison with the state of the art}
\label{sub:exp_compare}
To further evaluate the effectiveness of CTI, we conduct a detailed comparison with the current state of the art. For FFOE VQA, we compare our proposal with the recent state-of-the-art methods on TDIUC and VQA-2.0 datasets, including SAN \cite{Yang2016StackedAN}, QTA \cite{MTL_QTA}, BAN2 \cite{Kim2018BilinearAN}, Bottom-up \cite{tip-trick}, MCB \cite{Fukui2016MultimodalCB}, and RAU \cite{noh2016training}. For MC VQA, we compare with the state-of-the-art methods on Visual7W dataset, including BAN2 \cite{Kim2018BilinearAN}, SAN \cite{Yang2016StackedAN}, MLP 
\cite{jabri2016revisiting}, MCB 
\cite{Fukui2016MultimodalCB}, STL \cite{wang2018structuredSTL}, and fPMC 
\cite{hu2018learningfPMC}). It is worth noting that depending on tasks FFOE VQA or MC VQA, we use different representations for images and questions as clearly mentioned in Section \ref{sec:VQA_integrated}. This ensures a fair comparison with compared methods.

\begin{table}[!t]
\begin{center}
\small
\begin{tabular}{c|c|c|c|c|c}
\hline
\multirow{2}{*}{\textbf{Models}} & \multicolumn{5}{c}{\textbf{Evaluation metrics}}                                \\ \cline{2-6} 
                                 & \textbf{Acc}  & \textbf{Ari}  & \textbf{Har}  & \textbf{Ari-N} & \textbf{Har-N} \\ \hline
\textbf{BAN2 \cite{Kim2018BilinearAN}}                   & 85.5          & 67.4          & 54.9          & 37.4           & 15.7           \\ 
\textbf{SAN \cite{Yang2016StackedAN}}                   & 82.3          & 65.0            & 53.7          & 35.4           & 14.7           \\ 
\textbf{QTA \cite{MTL_QTA}}                     & 85.0          & 69.1          & 60.1          & \_             & \_             \\ 
\textbf{MCB \cite{Fukui2016MultimodalCB}}                   & 79.2          & 65.8          & 58.0          & 39.8           & 24.8           \\ 
\textbf{RAU \cite{noh2016training}}                     & 84.3          & 67.8          & 59.0          & 41.0           & 24.0           \\ \hline
\textbf{SAN-CTI}                  & 84.5          & 68.7          & 59.9          & 41.3           & 23.3           \\ 
\textbf{BAN2-CTI}                  & \textbf{87.0} & \textbf{72.5} & \textbf{65.5} & \textbf{45.8}  & \textbf{28.6}  \\ \hline
\end{tabular}
\end{center}
\caption{Performance comparison between different approaches with different evaluation metrics on TDIUC validation set. \textbf{BAN2-CTI} and \textbf{SAN-CTI} are the student models trained under our compact trilinear interaction teacher model.}
\label{tab:compare_app_tdiuc}
\end{table}

Regarding FFOE VQA, Tables \ref{tab:abl_vqa2} and   \ref{tab:compare_app_tdiuc} show comparative results on VQA-2.0 and TDIUC respectively. 
Specifcaly, Table \ref{tab:compare_app_tdiuc} shows that our distilled student BAN2-CTI outperforms all compared methods over all metrics by a large margin, i.e., our model outperforms the current  state-of-the-art QTA \cite{MTL_QTA} on TDIUC by $3.4\%$ and $5.4\%$ on \textit{Ari} and \textit{Har} metrics, respectively.
The results confirm that the proposed trilinear interaction has learned informative representations from the three inputs and the learned information is effectively transferred to student models by distillation.


\begin{table}[!t]
\begin{center}
\footnotesize
\begin{tabular}{|c|c|c|}
\hline
\multicolumn{1}{|c|}{\textbf{Dataset}}                                       & \textbf{\begin{tabular}[c]{@{}c@{}}Ref\\ models\end{tabular}} & \textbf{Acc-MC} \\ \hline
\multirow{8}{*}{\begin{tabular}[c]{@{}c@{}}\textbf{Visual7W}\\ \textbf{test set}\end{tabular}} & MLP \cite{jabri2016revisiting}                                                          & 67.1              \\ \cline{2-3} 
                                                                             
                                                                             & MCB \cite{Fukui2016MultimodalCB}                                                         & 62.2              \\ \cline{2-3} 
                                                                             & fPMC \cite{hu2018learningfPMC}                                                         & 66.0              \\ \cline{2-3} 
                             & STL \cite{wang2018structuredSTL}                                                           & 68.2              \\ \cline{2-3}                                                 & \multicolumn{1}{c|}{SAN \cite{Yang2016StackedAN}}                                      & 61.5              \\ \cline{2-3} 
                                                                             & \multicolumn{1}{c|}{BAN2 \cite{Kim2018BilinearAN}}                                      & 67.5              \\ \cline{2-3} 
                                         & \multicolumn{1}{c|}{\textbf{CTI}}                                      & \textbf{69.3}              \\ \cline{2-3}                                     & \multicolumn{1}{c|}{\textbf{CTIwBoxes}}                             & \textbf{72.3}              \\ \hline
\end{tabular}
\end{center}
\caption{Performance comparison between different approaches on Visual7W test set.  Both training set and validation set are used for training. All models but \textbf{CTIwBoxes} are trained with same image and question representations. Both train set and validation set are used for training. Note that \textbf{CTIwBoxes} is the proposed \textbf{CTI} model using Bottom-up features \cite{bottom-up2017} instead of grid features for image representation.
}
\label{tab:compare_app_v7w_vqa1}
\vspace{-0.25 cm}
\end{table}

Regarding MC VQA, 
Table  \ref{tab:compare_app_v7w_vqa1} shows that the proposed model (denoted as \textbf{CTI} in Table  \ref{tab:compare_app_v7w_vqa1}) outperforms compared methods by a noticeable margin. Our model  outperforms the current state-of-the-art STL \cite{wang2018structuredSTL} $1.1\%$. 
Again, this validates the effectiveness of the proposed joint presentation learning, which precisely and simultaneously learns interactions between the three inputs. 
We note that when comparing with other methods on Visual7W, for image representations, we used the grid features extracted from ResNet-512 \cite{he2016deepResnet152} for a fair comparison. Our proposed model can achieve further improvements by using the object detection-based features used in FFOE VQA. With new features, our model denoted as \textbf{CTIwBoxes} in Table \ref{tab:compare_app_v7w_vqa1} achieve $72.3\%$ accuracy with \textit{Acc-MC} metric which improves over the current state-of-the-art STL \cite{wang2018structuredSTL} $4.1\%$.

\subsection{Further analysis}
\label{sec:further_analysis}

\textbf{The effectiveness of PARALIND decomposition.} 
In this section, we compute the decomposition rate of PARALIND. For a fully interaction between the three inputs,  using (\ref{eq:hypothesis}), we would need to learn $2199.02$ billions parameters which is infeasible in practice. By using the PARALIND decomposition presented in Section \ref{Sec:proposal} with the provided settings, i.e., the number of slicing $R=32$ and the dimension of the joint representation $d_z=1024$, the number of parameters that need to learn is only $33.69$ millions. In the other words, we achieve a decomposition rate $\approx65,280$. 

\textbf{Compact Trilinear Interaction as the generalization of BAN \cite{Kim2018BilinearAN}.}
The proposed compact trilinear interaction model can be seen as a generalization of the state-of-the-art joint embedding BAN \cite{Kim2018BilinearAN}.
In BAN, each input contains an image representation $V$ $ \in \mathds{R}^{d_v \times v}$ 
and a question representation $Q \in \mathds{R}^{d_q \times q }$. 
The trilinear interaction model can be modified to adapt to these two inputs. The joint representation $z \in \mathds{R}^{d_z}$ in (\ref{eq:hypothesis}) can be adapted for two input as
\begin{equation}
z^T= \left(\mathcal{T}_{vq} \times_1 vec(V) \right) \times_2 vec(Q)
\label{eq:ban_hypothesis}
\end{equation}
where $\mathcal{T}_{vq} \in \mathds{R}^{d_V \times d_Q \times d_z}$ is a learnable tensor; $vec(V)$ is the vectorization of $V$ and $vec(Q)$  is the vectorization of $Q$ which output row vectors; $d_V = d_v \times v$; $d_Q = d_q \times q$.

By applying ``Parameter factorization" described in Section \ref{sub:factorization}, $z$ in (\ref{eq:ban_hypothesis}) can be approximated based on (\ref{eq:final_hypo_VQA}) as
\begin{equation}
z^T= \sum_{i=1}^{v}\sum_{j=1}^{q} \mathcal{M}_{ij}\left( V_{i}^{T}W_{z_v} \circ Q_{j}^{T}W_{z_q}\right)
\label{eq:decompose_BAN}
\end{equation}
where $W_{z_v} \in \mathds{R}^{d_v \times d_z}$ and $W_{z_q} \in \mathds{R}^{d_q \times d_z}$ are learnable factor matrices; $\mathcal{M}_{ij}$ is an attention weight of attention map $\mathcal{M} \in \mathds{R}^{v \times q}$ which can be computed from  (\ref{eq:tri_attmap_decomp_VQA}) as
\begin{equation}
\mathcal{M} = \sum^R_{r=1} {\llbracket  \mathcal{G}_r; V^{T}W_{v_r},  Q^{T}W_{q_r}\rrbracket}
\label{eq:tri_att_BAN}
\end{equation}
where $W_{v_r} \in \mathds{R}^{d_v \times d_{v_r}}$ and $W_{q_r} \in \mathds{R}^{d_q \times d_{q_r}}$ are learnable factor matrices; $d_{v_r} = d_v / R$; $d_{q_r} = d_q / R$; each $\mathcal{G}_r \in \mathds{R}^{d_{v_r} \times d_{q_r}}$ is a learnable Tucker tensor.

Interestingly, (\ref{eq:decompose_BAN}) can be reorganized to have a form of BAN~\cite{Kim2018BilinearAN} as
\begin{equation}
z_k= \sum_{i=1}^{v}\sum_{j=1}^{q} \mathcal{M}_{ij}\left( V_{i}^{T}\left(W_{{z_v}_k} W_{{z_q}_k}^{T}\right)Q_{j}\right)
\label{eq:final_hypo_VQA_BAN}
\end{equation}
where $z_k$ is the $k^{th}$ element of the joint representation $z$; $W_{{z_v}_k}$ and $W_{{z_q}_k}$ are $k^{th}$ column in factor matrices $W_{z_v}$ and $W_{z_q}$. 
Note that in (\ref{eq:final_hypo_VQA_BAN}), our attention map $\mathcal{M}$ is resulted from the PARALIND decomposition, while in BAN~\cite{Kim2018BilinearAN}, their attention map is computed by bilinear pooling.

%% file: conclusion.tex
\section{Conclusion}
We propose a novel compact trilinear interaction  which simultaneously learns high level associations between image, question, and answer in both MC VQA and FFOE VQA. In addition, knowledge distillation is the first time applied to FFOE VQA to overcome the computational complexity and memory issue of the interaction. The extensive experimental results show that the proposed models achieve the state-of-the-art results on three benchmarking datasets.  


%% file: main.bbl
\begin{thebibliography}{10}\itemsep=-1pt

\bibitem{2017AgrawalPriorVQA}
Devi~Parikh Aishwarya~Agrawal, Dhruv~Batra and Aniruddha Kembhavi.
\newblock Don't just assume; look and answer: Overcoming priors for visual
  question answering.
\newblock In {\em CVPR}, 2018.

\bibitem{bottom-up2017}
Peter Anderson, Xiaodong He, Chris Buehler, Damien Teney, Mark Johnson, Stephen
  Gould, and Lei Zhang.
\newblock Bottom-up and top-down attention for image captioning and {VQA}.
\newblock In {\em CVPR}, 2018.

\bibitem{VQA}
Stanislaw Antol, Aishwarya Agrawal, Jiasen Lu, Margaret Mitchell, Dhruv Batra,
  C.~Lawrence Zitnick, and Devi Parikh.
\newblock {VQA}: {V}isual {Q}uestion {A}nswering.
\newblock In {\em ICCV}, 2015.

\bibitem{ba2014deep}
Jimmy Ba and Rich Caruana.
\newblock Do deep nets really need to be deep?
\newblock In {\em NIPS}, 2014.

\bibitem{Benyounes2017MUTANMT}
Hedi Ben-younes, R{\'e}mi Cad{\`e}ne, Matthieu Cord, and Nicolas Thome.
\newblock Mutan: Multimodal tucker fusion for visual question answering.
\newblock In {\em ICCV}, 2017.

\bibitem{bro2009modelingPARALIND}
Rasmus Bro, Richard~A Harshman, Nicholas~D Sidiropoulos, and Margaret~E Lundy.
\newblock Modeling multi-way data with linearly dependent loadings.
\newblock {\em Journal of Chemometrics: A Journal of the Chemometrics Society},
  pages 324--340, 2009.

\bibitem{chen2017learning}
Guobin Chen, Wongun Choi, Xiang Yu, Tony Han, and Manmohan Chandraker.
\newblock Learning efficient object detection models with knowledge
  distillation.
\newblock In {\em NIPS}, 2017.

\bibitem{Fukui2016MultimodalCB}
Akira Fukui, Dong~Huk Park, Daylen Yang, Anna Rohrbach, Trevor Darrell, and
  Marcus Rohrbach.
\newblock Multimodal compact bilinear pooling for visual question answering and
  visual grounding.
\newblock In {\em EMNLP}, 2016.

\bibitem{gan2017vqs}
Chuang Gan, Yandong Li, Haoxiang Li, Chen Sun, and Boqing Gong.
\newblock Vqs: Linking segmentations to questions and answers for supervised
  attention in vqa and question-focused semantic segmentation.
\newblock In {\em ICCV}, 2017.

\bibitem{vqav22016}
Yash Goyal, Tejas Khot, Douglas Summers{-}Stay, Dhruv Batra, and Devi Parikh.
\newblock Making the {V} in {VQA} matter: Elevating the role of image
  understanding in visual question answering.
\newblock In {\em CVPR}, 2017.

\bibitem{gupta2016cross}
Saurabh Gupta, Judy Hoffman, and Jitendra Malik.
\newblock Cross modal distillation for supervision transfer.
\newblock In {\em CVPR}, 2016.

\bibitem{he2016deepResnet152}
Kaiming He, Xiangyu Zhang, Shaoqing Ren, and Jian Sun.
\newblock Deep residual learning for image recognition.
\newblock In {\em CVPR}, 2016.

\bibitem{hinton2015distilling}
Geoffrey Hinton, Oriol Vinyals, and Jeff Dean.
\newblock Distilling the knowledge in a neural network.
\newblock In {\em NIPS Deep Learning Workshop}, 2014.

\bibitem{hu2018learningfPMC}
Hexiang Hu, Wei-Lun Chao, and Fei Sha.
\newblock Learning answer embeddings for visual question answering.
\newblock In {\em CVPR}, 2018.

\bibitem{ilievski2016focused}
Ilija Ilievski, Shuicheng Yan, and Jiashi Feng.
\newblock A focused dynamic attention model for visual question answering.
\newblock {\em arXiv preprint arXiv:1604.01485}, 2016.

\bibitem{jabri2016revisiting}
Allan Jabri, Armand Joulin, and Laurens Van Der~Maaten.
\newblock Revisiting visual question answering baselines.
\newblock In {\em ECCV}, 2016.

\bibitem{Kushal2018Tdiuc}
Kushal Kafle and Christopher Kanan.
\newblock An analysis of visual question answering algorithms.
\newblock In {\em ICCV}, 2017.

\bibitem{Kim2018BilinearAN}
Jin-Hwa Kim, Jaehyun Jun, and Byoung-Tak Zhang.
\newblock Bilinear attention networks.
\newblock In {\em NIPS}, 2018.

\bibitem{kim2016multimodal}
Jin-Hwa Kim, Sang-Woo Lee, Donghyun Kwak, Min-Oh Heo, Jeonghee Kim, Jung-Woo
  Ha, and Byoung-Tak Zhang.
\newblock Multimodal residual learning for visual qa.
\newblock In {\em NIPS}, 2016.

\bibitem{kim2016hadamard}
Jin-Hwa Kim, Kyoung-Woon On, Woosang Lim, Jeonghee Kim, Jung-Woo Ha, and
  Byoung-Tak Zhang.
\newblock Hadamard product for low-rank bilinear pooling.
\newblock In {\em ICLR}, 2017.

\bibitem{kolda2009tensor}
Tamara~G Kolda and Brett~W Bader.
\newblock Tensor decompositions and applications.
\newblock {\em SIAM review}, pages 455--500, 2009.

\bibitem{2016LinFPN}
Tsung-Yi Lin, Piotr Doll{\'a}r, Ross~B. Girshick, Kaiming He, Bharath
  Hariharan, and Serge~J. Belongie.
\newblock Feature pyramid networks for object detection.
\newblock In {\em CVPR}, 2017.

\bibitem{lu2016hierarchical}
Jiasen Lu, Jianwei Yang, Dhruv Batra, and Devi Parikh.
\newblock Hierarchical question-image co-attention for visual question
  answering.
\newblock In {\em NIPS}, 2016.

\bibitem{Ma2017VisualQA}
Chao Ma, Chunhua Shen, Anthony~R. Dick, and Anton van~den Hengel.
\newblock Visual question answering with memory-augmented networks.
\newblock In {\em CVPR}, 2018.

\bibitem{malinowski2014towards}
Mateusz Malinowski and Mario Fritz.
\newblock Towards a visual turing challenge.
\newblock In {\em NIPS workshop}, 2014.

\bibitem{Malinowski2015AskYN}
Mateusz Malinowski, Marcus Rohrbach, and Mario Fritz.
\newblock Ask your neurons: A neural-based approach to answering questions
  about images.
\newblock {\em ICCV}, pages 1--9, 2015.

\bibitem{nam2017dual}
Hyeonseob Nam, Jung-Woo Ha, and Jeonghee Kim.
\newblock Dual attention networks for multimodal reasoning and matching.
\newblock In {\em CVPR}, 2017.

\bibitem{dense-attention}
Duy-Kien Nguyen and Takayuki Okatani.
\newblock Improved fusion of visual and language representations by dense
  symmetric co-attention for visual question answering.
\newblock In {\em CVPR}, 2018.

\bibitem{noh2016training}
Hyeonwoo Noh and Bohyung Han.
\newblock Training recurrent answering units with joint loss minimization for
  vqa.
\newblock {\em arXiv preprint arXiv:1606.03647}, 2016.

\bibitem{noh2016image}
Hyeonwoo Noh, Paul Hongsuck~Seo, and Bohyung Han.
\newblock Image question answering using convolutional neural network with
  dynamic parameter prediction.
\newblock In {\em CVPR}, 2016.

\bibitem{paszke2017automaticPyTorch}
Adam Paszke, Sam Gross, Soumith Chintala, Gregory Chanan, Edward Yang, Zachary
  DeVito, Zeming Lin, Alban Desmaison, Luca Antiga, and Adam Lerer.
\newblock Automatic differentiation in pytorch.
\newblock In {\em NIPS 2017 Workshop}, 2017.

\bibitem{pennington2014glove}
Jeffrey Pennington, Richard Socher, and Christopher~D. Manning.
\newblock Glove: Global vectors for word representation.
\newblock In {\em EMNLP}, 2014.

\bibitem{romero2014fitnets}
Adriana Romero, Nicolas Ballas, Samira~Ebrahimi Kahou, Antoine Chassang, Carlo
  Gatta, and Yoshua Bengio.
\newblock Fitnets: Hints for thin deep nets.
\newblock In {\em ICLR}, 2015.

\bibitem{schwartz2017high}
Idan Schwartz, Alexander Schwing, and Tamir Hazan.
\newblock High-order attention models for visual question answering.
\newblock In {\em NIPS}, 2017.

\bibitem{MTL_QTA}
Yang Shi, Tommaso Furlanello, Sheng Zha, and Animashree Anandkumar.
\newblock Question type guided attention in visual question answering.
\newblock In {\em ECCV}, 2018.

\bibitem{shih2016look}
Kevin~J Shih, Saurabh Singh, and Derek Hoiem.
\newblock Where to look: Focus regions for visual question answering.
\newblock In {\em CVPR}, 2016.

\bibitem{tip-trick}
Damien Teney, Peter Anderson, Xiaodong He, and Anton van~den Hengel.
\newblock Tips and tricks for visual question answering: Learnings from the
  2017 challenge.
\newblock In {\em CVPR}, 2018.

\bibitem{teney2016zero}
Damien Teney and Anton van~den Hengel.
\newblock Zero-shot visual question answering.
\newblock {\em arXiv preprint arXiv:1611.05546}, 2016.

\bibitem{DBLP:conf/cvpr/TeneyLH17}
Damien Teney, Lingqiao Liu, and Anton van~den Hengel.
\newblock Graph-structured representations for visual question answering.
\newblock In {\em CVPR}, 2017.

\bibitem{Teney2017VisualQA}
Damien Teney and Anton van~den Hengel.
\newblock Visual question answering as a meta learning task.
\newblock In {\em ECCV}, 2018.

\bibitem{wang2018structuredSTL}
Zhe Wang, Xiaoyi Liu, Limin Wang, Yu Qiao, Xiaohui Xie, and Charless Fowlkes.
\newblock Structured triplet learning with pos-tag guided attention for visual
  question answering.
\newblock In {\em WACV}, 2018.

\bibitem{xu2016ask}
Huijuan Xu and Kate Saenko.
\newblock Ask, attend and answer: Exploring question-guided spatial attention
  for visual question answering.
\newblock In {\em ECCV}, 2016.

\bibitem{Yang2016StackedAN}
Zichao Yang, Xiaodong He, Jianfeng Gao, Li Deng, and Alexander~J. Smola.
\newblock Stacked attention networks for image question answering.
\newblock In {\em CVPR}, 2016.

\bibitem{yu2017multi}
Dongfei Yu, Jianlong Fu, Tao Mei, and Yong Rui.
\newblock Multi-level attention networks for visual question answering.
\newblock In {\em CVPR}, 2017.

\bibitem{DBLP:conf/iccv/YuY0T17}
Zhou Yu, Jun Yu, Jianping Fan, and Dacheng Tao.
\newblock Multi-modal factorized bilinear pooling with co-attention learning
  for visual question answering.
\newblock In {\em ICCV}, 2017.

\bibitem{zhou2015simple}
Bolei Zhou, Yuandong Tian, Sainbayar Sukhbaatar, Arthur Szlam, and Rob Fergus.
\newblock Simple baseline for visual question answering.
\newblock {\em arXiv preprint arXiv:1512.02167}, 2015.

\bibitem{zhu2016visual7w}
Yuke Zhu, Oliver Groth, Michael Bernstein, and Li Fei-Fei.
\newblock Visual7{W}: Grounded question answering in images.
\newblock In {\em CVPR}, 2016.

\end{thebibliography}
